\documentclass[10pt,twocolumn,letterpaper]{article}
\pdfoutput=1
\usepackage{iccv}
\usepackage{times}
\usepackage{epsfig}
\usepackage{graphicx}
\usepackage{amsmath}
\usepackage{amssymb}

\usepackage{multirow}
\usepackage{subcaption}
\usepackage{algorithm}
\usepackage{algorithmic}

\usepackage[breaklinks=true,bookmarks=false]{hyperref}

\iccvfinalcopy 

\def\iccvPaperID{****} 

\begin{document}

\title{Progressive Multi-scale Fusion Network for RGB-D Salient Object Detection}

\author{Guangyu Ren\\
Imperial College London\\
{\tt\small g.ren19@imperial.ac.uk}

\and
Yanchun Xie\\
University of Liverpool\\
{\tt\small xyc2690@163.com}

\and
Tianhong Dai\\
Imperial College London\\
{\tt\small tianhong.dai15@imperial.ac.uk}

\and
Tania Stathaki\\
Imperial College London\\
{\tt\small t.stathaki@imperial.ac.uk}
}

\maketitle
\begin{abstract}
  Salient object detection(SOD) aims at locating the most significant object within a given image. In recent years, great progress has been made in applying SOD on many vision tasks. The depth map could provide additional spatial prior and boundary cues to boost the performance. Combining the depth information with image data obtained from standard visual cameras has been widely used in recent SOD works, however, introducing depth information in a suboptimal fusion strategy may have negative influence in the performance of SOD. In this paper, we discuss about the advantages of the so-called progressive multi-scale fusion method and propose a mask-guided feature aggregation module(MGFA). The proposed framework can effectively combine the two features of different modalities and, furthermore, alleviate the impact of erroneous depth features, which are inevitably caused by the variation of depth quality. We further introduce a mask-guided refinement module(MGRM) to complement the high-level semantic features and reduce the irrelevant features from multi-scale fusion, leading to an overall refinement of detection. Experiments on five challenging benchmarks demonstrate that the proposed method outperforms 11 state-of-the-art methods under different evaluation metrics.
\end{abstract}

\section{Introduction}
Salient object detection(SOD) aims at detecting prominent and important objects in a given image under consideration. 
SOD is a crucial part of numerous computer vision tasks and has been applied in many different fields, such as semantic segmentation~\cite{wang2018weaklysegmentation,wei2017objectsegmentation}, scene classification~\cite{ren2013regionclassification}, person re-identification~\cite{zhao2013unsupervisedreid}, visual tracking~\cite{mahadevan2009saliencytracking}, video summarization~\cite{simakov2008videosummarizing} and others.
The significance of SOD together with the rapid development of deep learning methods has led to substantial progress in recent years. However, there are still challenging issues on SOD, especially in complex scenarios where salient objects are placed on cluttered backgrounds. 

\begin{figure}
\minipage{0.15\textwidth}
  \centering
  \includegraphics[width=\linewidth]{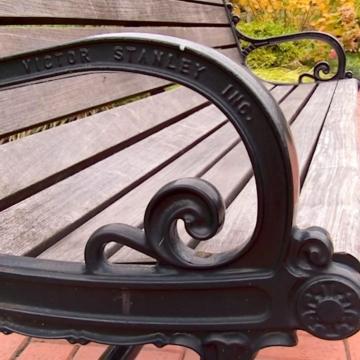}
  (a) Image
\endminipage\hfill
\minipage{0.15\textwidth}
  \centering
  \includegraphics[width=\linewidth]{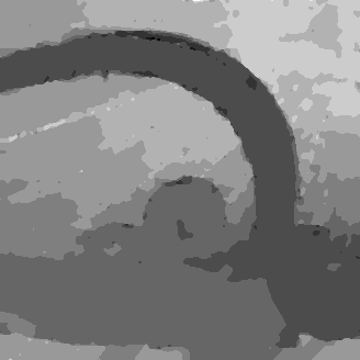}
  (b) Depth
\endminipage\hfill
\minipage{0.15\textwidth}
  \centering
  \includegraphics[width=\linewidth]{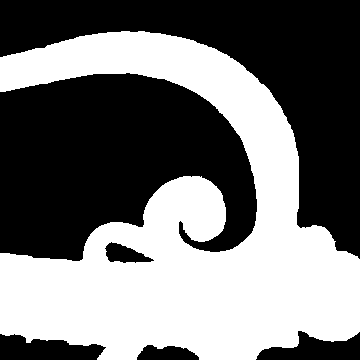}
  (c) GT
\endminipage\hfill
\minipage{0.15\textwidth}
  \centering
  \includegraphics[width=\linewidth]{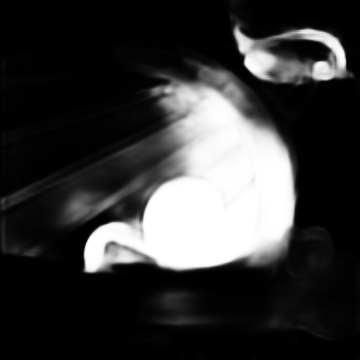}
  (d) RGB based
\endminipage\hfill
\minipage{0.15\textwidth}
  \centering
  \includegraphics[width=\linewidth]{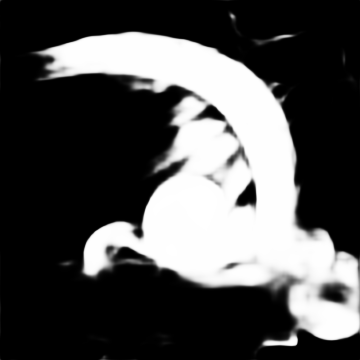}
  (e) RGB-D based
\endminipage\hfill
\minipage{0.15\textwidth}
  \centering
  \includegraphics[width=\linewidth]{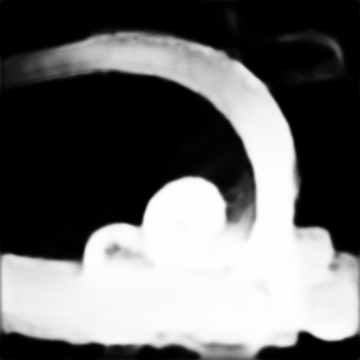}
  (f) Ours
\endminipage\hfill

\caption{Qualitative visual results of RGB and RGB-D based methods. (a) original image, (b) depth image, (c) the ground truth, (d) PoolNet~\cite{liu2019poolnet},(e) CPFP~\cite{zhao2019contrast}, (f) our proposed method.}
\label{introdcution}
\end{figure}

\begin{figure*}
  \includegraphics[width=\textwidth]{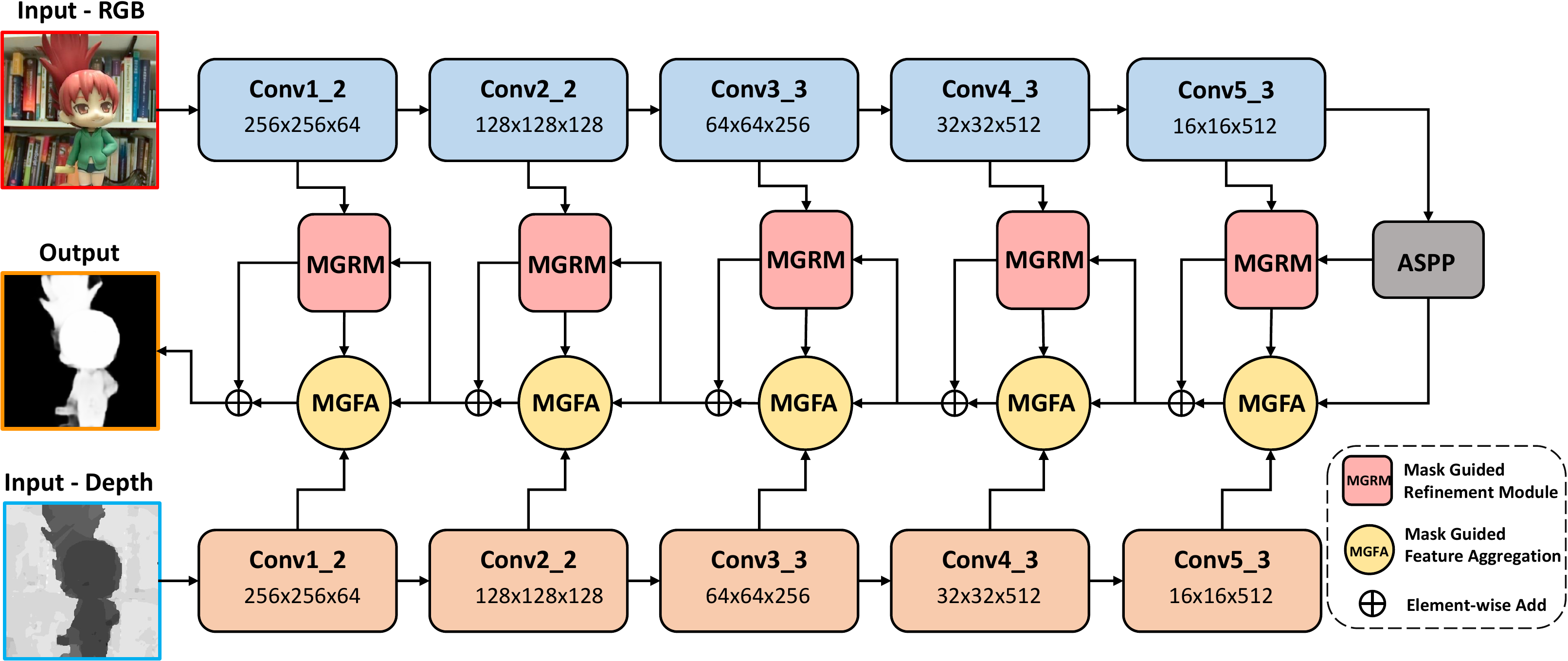}
  \caption{The overall structure of our proposed network, where RGB and depth images are fed into two VGG19 backbones and aggregated progressively from deep to shallow layers. ASPP is the atrous spatial pyramid pooling module\cite{chen2017aspp}.}
  \label{overall}
\end{figure*}

Depth maps can provide effective spatial information for salient object detection in complex scenes. Depth maps were firstly introduced in \cite{lang2012depth} to provide complementary information to further enhance the detection performance. In \cite{lang2012depth} the depth information is provided by the Kinect camera. The importance of depth information is illustrated in a representative example depicted in Figure \ref{introdcution}. It is seen that RGB-only based method~\cite{liu2019poolnet} exhibits unsatisfactory prediction results in complex scenes. Compared to Figure \ref{introdcution}(d), both (e) and (f) show more detail in the detection saliency map due to the additional depth information. However, erroneous depth information may cause negative impact on SOD. Zhao~\cite{zhao2019contrast} proposes a cross-modal fusion strategy to overcome this problem by enhancing the contrast between foreground and background objects in depth images. As shown in Figure \ref{introdcution}(e), in the joint method proposed in \cite{zhao2019contrast}, more detail is detected in the saliency map compared to RGB-only based method. However, the salient object is still incomplete and non-salient parts are also shown in the saliency map. This result indicates that the quality of depth images is uncertain and depends on the sensor. This observation implies that when we combine image data obtained from visual cameras with depth sensor data, the latter does not necessarily contribute positively to the final outcome of the particular algorithm, due to possible low quality of the depth sensor data.

Recent works \cite{peng2014rgbd,ren2015exploiting,song2017depth,liu2019salient,zhao2020single} adopt an early fusion strategy, which indicates that the features from different modalities are combined in shallow layers. In these works, common ways for cross-modal combination are concatenating the depth features with the original image features to form a four-channel input before feeding it into the network. This approach could potentially damage the robust RGB features in case concatenation involves low-quality depth maps.
Besides, using solely early fusion strategies cannot generate satisfactory prediction results as demonstrated in \cite{zhao2020single}. The additional depth information should be fed into the network only if it results in further boosting the detection performance.

To tackle the above issues, we seek for alternative fusion strategies. We consider the RGB-Depth feature fusion in a progressive way and propose a novel network with a Mask-Guided Feature Aggregation (MGFA) module, which can take advantages of multi-scale features and effectively introduce useful depth information.

However, multi-scale fusion suffers from the following drawbacks. Firstly, introducing redundant information can impede performance. The redundant features will lead to the emergence of objects unrelated to the target object in the final predictions, which could greatly affect the accuracy of the output saliency map. Pang in \cite{pang2020multi} investigates the multi-scale issue and aims at gaining benefits from multi-scale features. More, specifically, the authors in~\cite{pang2020multi} design a transformation-interaction-fusion strategy and propose an aggregate interaction module to utilize the features from adjacent layers, which could introduce less noise due to small sampling rates. However, this method utilizes extra parameters on mutual learning and self-interaction. The second drawback is that high-level features can be gradually diluted from deep to shallow layers. As demonstrated in \cite{ren2020salient,liu2019poolnet}, high-level semantic features are crucial for salient object detection due to their paramount importance among the abundant semantic features. The authors adopt a feature guidance strategy to complement the high-level semantic features, which can remedy the loss of semantic features when they pass from deep to shallow layers in the feature pyramid network(FPN)~\cite{lin2017feature}. Different from the above-mentioned methods, considering though the same issues, we design a novel refinement module to filter the noise and irrelevant information and further complement semantic information in an effective way.

Our main contributions can be concluded as follows:
\begin{itemize}
\item We construct a two-stream network with a mask-guided feature aggregation module to extract RGB features and the depth features. The proposed network considers merging the two-modal features from deep layers to shallow layers in a progressive way and the mask-guided feature aggregation module can effectively merge two-modal features by the guidance of deep-layer features, leading to the alleviation of the negative impact of unstable depth maps.

\item We further design a mask-guided refinement module to diminish the noise features in the multi-scale setting, semantic features can be maintained well due to introducing deep-layer features and the noise can be filtered out efficiently with the guide of the prediction mask.

\item We evaluate our proposed method with 11 state-of-the-art SOD methods on five benchmarks. It achieves competitive performance against previous methods with respect to different evaluation metrics.
\end{itemize}

\section{Related Work}

\subsection{RGB Salient Object Detection}
The significance of salient object detection together with the rapid development of deep learning has led to substantial progress in recent years. Wei~\cite{wei2020label} investigates pixels at different locations and proposes a label decoupling procedure to separate a label into a body map and a detailed map. The authors in \cite{wei2020label} use a feature interaction network to learn the features between branches. Ren~\cite{ren2020salient} proposes a pyramid self-attention module to enlarge the receptive field of the network and to further complement the high-level semantic features to the top-down path of the FPN. Liu~\cite{liu2019poolnet} designs a simple pooling-based network for SOD tasks. They adopt a global guidance module to guide high-level features and a feature aggregation module to combine the coarse and fine features. However, RGB-based SOD may generate unsatisfactory saliency maps in a complex environment.

\begin{figure}
  \includegraphics[width=\linewidth]{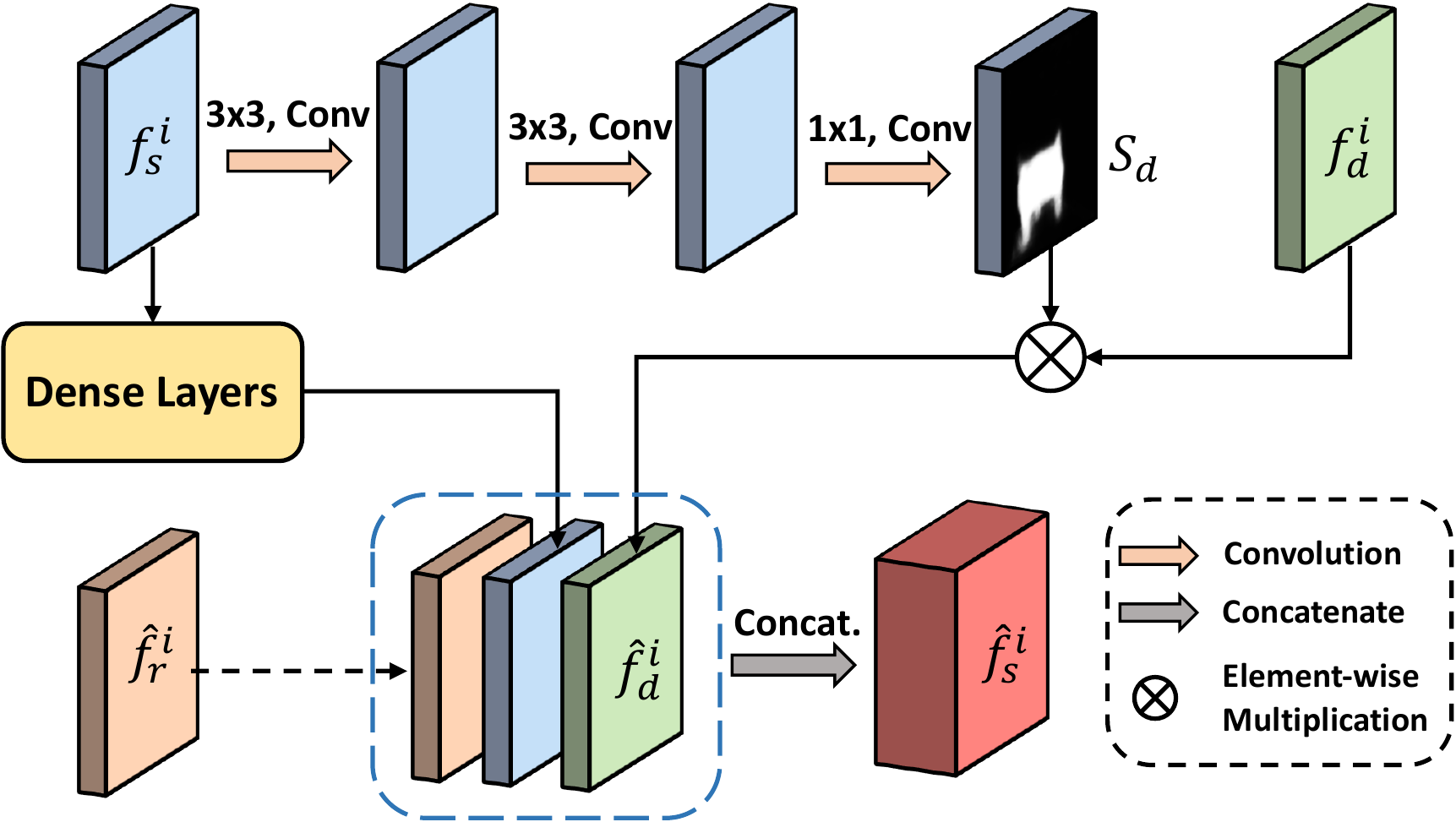}
  \caption{The proposed mask-guided feature aggregation. $\hat{f^{i}_{r}}$ and ${f^{i}_{d}}$ represent the $i$-th level of RGB and depth features, respectively. $f^{i}_{s}$ is the output of previous MGFA ${\hat{f}^{i+1}_{s}}$.}
  \label{maskguided}
\end{figure}

\subsection{RGB-D Salient Object Detection}
Over the past few years, traditional models~\cite{cong2016saliency,guo2016salient,ciptadi2013depth} utilize jointly the useful attributes from RGB and depth cues to design handcrafted features. However, the use of such type of features limits the performance of traditional methods. More specifically, handcrafted features are normally inconsistent and have low representation ability, leading to unsatisfactory detection results, especially in complex scenarios.

In recent years, deep learning models are adopted to fuse RGB and depth data since fully convolutional networks show superiority on SOD due to their ability in extracting robust features. 
Zhao~\cite{zhao2019contrast} utilizes contrast priors to enhance the contrast between foreground and background objects in depth maps. Zhao~\cite{zhao2020single} designs a single stream network and adopts both early fusion and middle fusion strategies. This method learns the cross-modal interaction in the encoder and further enhances its effect in the decoder. Pang~\cite{pang2020hierarchical} proposes a dense connected structure to integrate cross-modal features, which can produce region-aware dynamic filters to guide the decoder. Chen~\cite{chen2020progressively} treats depth images as mid-level or high-level feature maps and adopts a lightweight network to obtain depth features. Besides, the authors in~\cite{chen2020progressively} propose an alternate refinement strategy in a coarse-to-fine manner to alternately use RGB and depth features. Piao~\cite{piao2020a2dele} investigates adaptive and attention distillation schemes and connects two modalities by designing a depth distiller, which can transfer reliable and avoid erroneous depth information.
Different from the aforementioned methods, our work aims to utilize a mask-guided mechanism to alleviate the impact of noise which is caused by low-quality depth features and multi-scale information.

\section{The Proposed Framework}
In this section, we first introduce the proposed overall architecture and then present two main components in detail, namely, the proposed novel Mask-Guided Feature Aggregation module(MGFA) and the Mask-Guided Refinement Module(MGRM).

\subsection{The Overall Architecture}
We propose a two-stream network which is illustrated in Figure \ref{overall}. This structure follows a coarse-to-fine network structure and has two inputs, namely, RGB images and depth images, which are fed into RGB stream and depth stream respectively. The two streams have the same feature extraction structure. Both of them utilize the 19-layer deep convolutional neural network VGG19~\cite{simonyan2014vgg} as a backbone to generate multi-level features with 5 different resolutions, which can be denoted as $\left\{f^1_{r},f^2_{r},f^3_{r},f^4_{r},f^5_{r}\right\}$ and $\left\{f^1_{d},f^2_{d},f^3_{d},f^4_{d},f^5_{d}\right\}$. More specifically, instead of directly combining them, we input an initial prediction map which is generated by a semantic segmentation module which follows the Atrous Spatial Pyramid Pooling (ASPP) approach~\cite{chen2017aspp} with RGB and depth features into our proposed MGFA to aggregate two-modal features in the corresponding scales. Then, the aggregated output is fed into another MGFA which treats the output as a saliency map. Furthermore, an MGRM is applied to refine the RGB features by inputting the same saliency map in each scale.

\begin{figure}[t]
  \includegraphics[width=\linewidth]{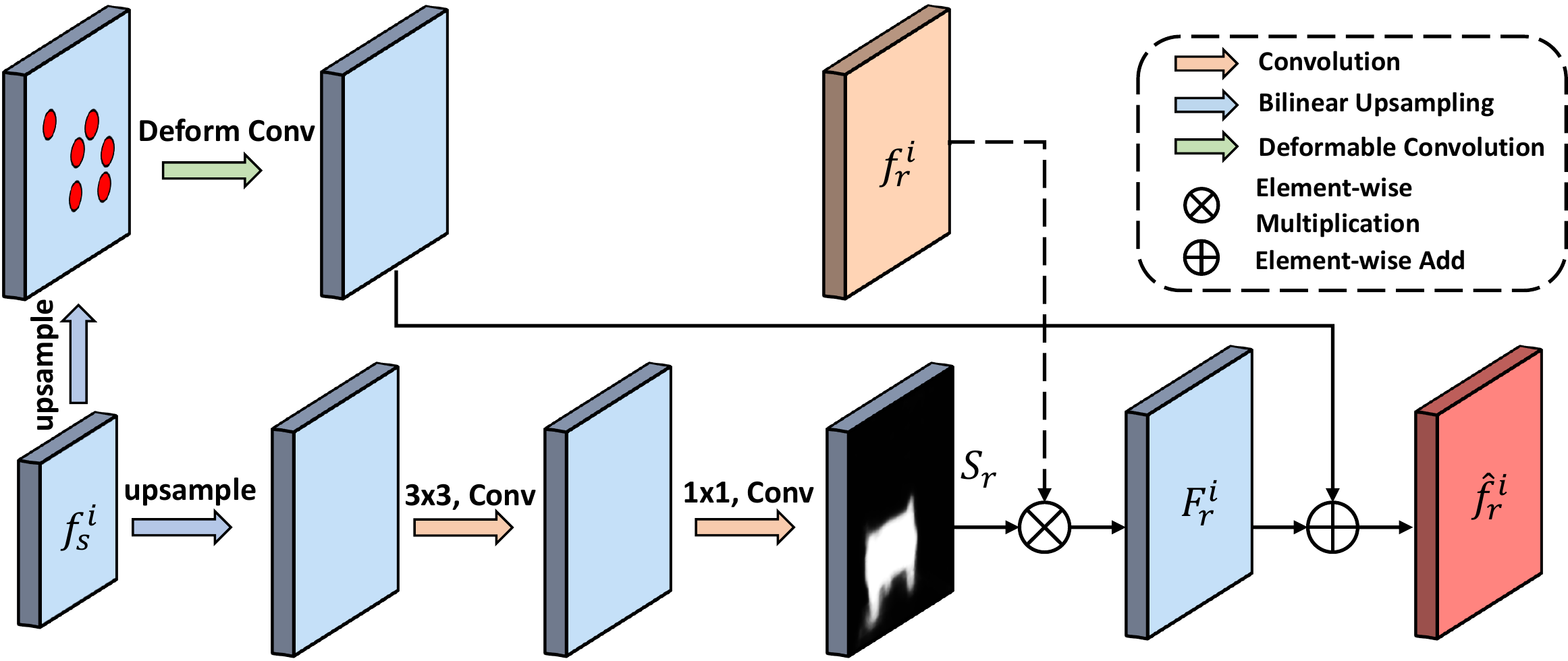}
  \caption{The proposed mask-guided refinement module.}
  \label{maskrefinement}
\end{figure}

\subsection{Mask-Guided Feature Aggregation}
As mentioned above, the early fusion approach could potentially damage the robust RGB features due to low-quality depth maps and the detection performance also depends on the quality of depth maps.
In order to tackle these issues, we propose, as mentioned, a novel Mask-Guided Feature Aggregation (MGFA) module which aims to take full merit of both RGB and depth features. As shown in Figure \ref{maskguided}, the initial input saliency map of MGFA is generated by the final stage of RGB stream which is termed as $f_{s}$. Inspired by \cite{pang2019maskguided}, the mask $S_{d}$ is obtained as follows:

\begin{equation}
    \centering
     S_{d} = \sigma(N_{1}(f_{s}))
\end{equation}
where $N_{1}$ indicates two 3x3 convolution layers and a 1x1 convolution layer. $\sigma$ refers to the sigmoid function. 
Then, the filtered depth feature map denoted by $\hat{f_{d}}$ is obtained by applying element-wise product between $S_{d}$ and $f_{d}$ as follows:

\begin{equation}
    \centering
    \hat{f_{d}} = S_{d}\cdot f_{d}
\end{equation}
After processed by the mask-guided operation, the unreliable features from low-quality depth images are filtered and then we use a dense layer~\cite{huang2017densely} to obtain a more powerful feature map $S_{dense}$ with abundant receptive fields and information details. The integrated feature map $\hat{f_{s}}$ is generated by the different modal features, namely, $S_{dense}$, $\hat{f_{d}}$ and $\hat{f_{r}}$, where $\hat{f_{r}}$ is the refined RGB feature map described in section 3.3. 
\begin{equation}
    \centering
    \hat{f_{s}} = Cat(\hat{f_{r}},\hat{f_{d}},S_{dense})
\end{equation}
where $Cat$ represents the concatenation operation. Finally, the output $\hat{f_{s}}$ is fed into the next MGFA module with $\hat{f_{r}}$ to guide the shallower layer’s depth information and the ground truth is applied to further supervise the $i$-th output.

This progressive feature aggregation strategy has the following benefits. First, transferring two-modal information into multi-scale features and aggregate them progressively from deep to shallow can overcome the drawbacks. Specifically, the generated mask can effectively filter the noise from depth features, especially within shallow levels, leading to alleviate the impact of low-quality depth images. Furthermore, along with the RGB and depth information, the output from the previous deeper level is provided to further refine the feature aggregation and complement the semantic features.

\subsection{Mask-Guided Refinement Module}
Multi-scale fusion strategy combines complementary information and global context information from both RGB and depth modalities. However, it also introduces irrelevant information which impedes the performance of salient object detection. As shown in Figure \ref{introdcution}(d) and (e), not only the salient object is detected, some irrelevant parts are also presented. To tackle this problem, we propose a Mask-Guided Refinement Module(MGRM) to filter the noise and irrelevant information in multiple scales. As illustrated in Figure \ref{maskrefinement}, sharing the similar mask-guided idea to MGFA, the RGB feature maps can be refined by the deeper-layer feature maps:

\begin{equation}
    \centering
    S_{r}  = \sigma(N_{2}(f_{s}))
\end{equation}
\begin{equation}
    \centering
    F_{r} = S_{r}\cdot f_{r}
\end{equation}
where $N_{2}$ indicates a 3x3 convolution layer and a 1x1 convolution layer. $S_{r}$ indicates the mask for RGB features and $F_{r}$ represents the filtered RGB features.
MGRM also diversifies the receptive fields of multiple scales. Specifically, the receptive field of RGB features in each scale can be enlarged by introducing the features from deeper layers:

\begin{equation}
    \centering
    \hat{f_{r}} = deformable(Up(f_{s}, f_{r})) + F_{r}
\end{equation}
where $Up$ represents bilinear interpolation operation to resize the deeper-layer feature map $f_{s}$ to the same size with $f_{r}$. $deformable$ refers to a deformable convolution layer with a flexible kernel to extract $f_{s}$ after upsampling. Finally, the filtered RGB $F_{r}$ is added to generate the output $\hat{f_{r}}$.

\begin{figure}[t!]
  \centering
  \includegraphics[width=\linewidth]{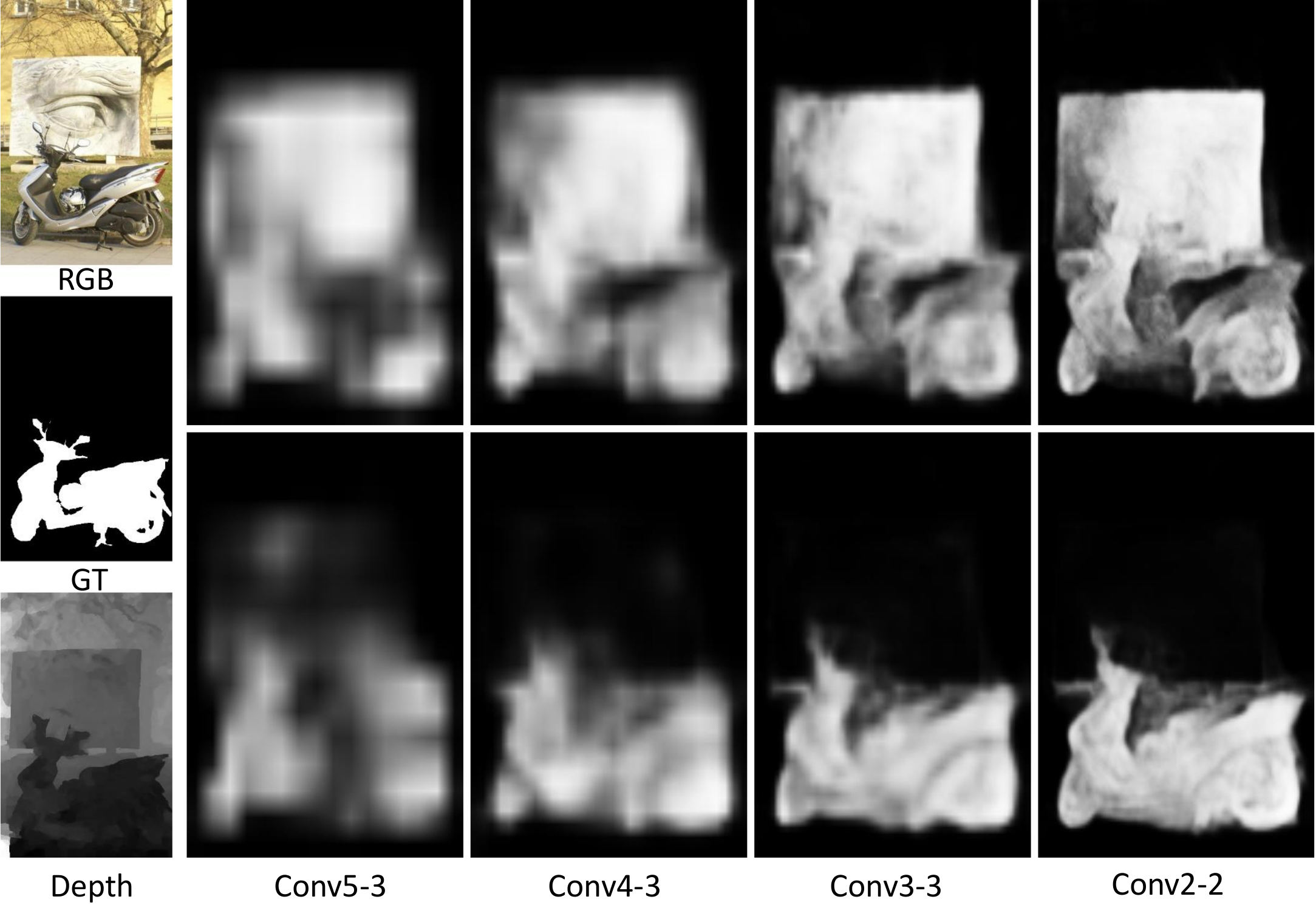}
\caption{Predictions from four different levels of the proposed network. $Conv5-3$ represents the saliency map which is generated by $f^5_{r}$ and $f^5_{d}$. Saliency maps in the first row are generated solely by MGFA whereas the bottom row maps are produced by MGFA and MGRM.}
\label{stage_comparison}
\end{figure}

Figure \ref{stage_comparison} shows the effectiveness of MGRM on different scales. The first row refers to the four different scales’ outputs without the refinement of MGRM and the bottom row represents the corresponding outputs with the MGRM. Obviously, the drawing board displayed in the background misleads the model to detect this non-salient object due to the low contrast of the RGB map. Without the MGRM, noise is also introduced with the global context information during multi-scale fusion. Compared to the erroneous phenomena described above, it is observed that MGRM can effectively introduce semantic features and filter the irrelevant features, leading to further detailed refinement of the saliency map. The proposed progressive fusion strategy is illustrated in algorithm \ref{algo:alg1}

\begin{algorithm}
	\caption{\label{alg:all} Progressive Multi-scale Fusion.
	}
	\begin{algorithmic}[1]
		\REQUIRE
		$f^{i}_{r}$ is the $i$-th level RGB features,
		$f^{i}_{d}$ is the $i$-th level depth features,
		$f^{i}_{s}$ is the deep semantic features.
		\FOR{i = 5, 4, 3, 2, 1}
		\STATE $S^{i}_{r}$ = Mask($f^{i}_{s}$) \hfill{$\triangleright$ Stage 1};
		\STATE $F^{i}_{r}$ = $S^{i}_{r} \cdot f^{i}_{r}$;
		\STATE $\hat{f}^{i}_{r}$ = $deformable(Up(f^{i}_{s}, f^{i}_{r})) + F^{i}_{r}$;
		\STATE $S^{i}_{d}$ = Mask($f^{i}_{s}$) \hfill{$\triangleright$ Stage 2};
		\STATE $\hat{f}^{i}_{d}$ = $S^{i}_{d} \cdot f^{i}_{d}$;
		\STATE $S^{i}_{dense}$ = Dense($f^{i}_{s}$);
		\STATE $\hat{f}^{i}_{s}$ = $Cat(\hat{f}^{i}_{r}, \hat{f}^{i}_{d}, S^{i}_{dense})$;
		\STATE ${Output}^{i}$ = $\hat{f}^{i}_{s} + \hat{f}^{i}_{r}$ \hfill{$\triangleright$ Stage 3};
		\IF{i != 1}
		\STATE $f^{i-1}_{s} = {Output}^{i}$;
		\ENDIF
		\ENDFOR
        \RETURN ${Output}^{1}$;
	\end{algorithmic}
	\label{algo:alg1}
\end{algorithm}

\section{Experiments}

\begin{table*}[h]
\centering
\caption{Quantitative comparison including the maximum of $F_{\beta}$, $S_{\alpha}$, $E_{\theta}$, and $M$, over five widely evaluated datasets. $\uparrow \& \downarrow$ represent higher and lower is better, respectively. The best two scores are highlighted in {\color{red} \textbf{red}} and {\color{blue} \textbf{blue}}.}
\label{table:qua}
\resizebox{\linewidth}{!}{
\begin{tabular}{cc|ccccccccccc|c}
\hline
\hline
\multirow{2}{*}{} & \multirow{2}{*}{Metric} & CDCP   & DF  & CTMF   & AFNet    & MMCI  & TANet  & DMRA   & CPFP   & D3Net  & A2dele  & DANet    & Ours  \\
                         &  & \cite{zhu2017innovative} & \cite{qu2017df}   & \cite{han2017ctmf}  & \cite{wang2019afnet}     & \cite{chen2019mmci}    & \cite{chen2019tanet}  & \cite{piao2019dmra} & \cite{zhao2019contrast}  & \cite{DBLP:journals/D3/abs-1907-06781} & \cite{piao2020a2dele}&  \cite{zhao2020single}    & $*$  \\ \hline
\multirow{4}{*}{\rotatebox{90}{NJUD}~\rotatebox{90}{~\cite{ju2015njud}}}    & $F_{\beta}\uparrow$ & 0.661 & 0.789 & 0.857 & 0.804 & 0.868 & 0.888 & 0.896 & 0.890 &0.903 &0.890  & {\textcolor{blue} {\textbf{0.910}}} & {\textcolor{red} {\textbf{0.919}}} \\
                         & $S_{\alpha}\uparrow$ & 0.672 & 0.735 & 0.849 & 0.772 & 0.859 & 0.878 & 0.885 & 0.878 & 0.895 &0.867 & {\textcolor{blue} {\textbf{0.899}}}  & {\textcolor{red} {\textbf{0.910}}} \\
                         & $E_{\theta}\uparrow$ & 0.751 & 0.818 & 0.866 & 0.847 & 0.882 & 0.909 & 0.920 & 0.900 & 0.901 &0.914 & {\textcolor{blue} {\textbf{0.922}}}  & {\textcolor{red} {\textbf{0.926}}} \\
                         & $M\,\downarrow$ & 0.182 & 0.151 & 0.085 & 0.100 & 0.079 & 0.061 & 0.051 & 0.053 & 0.051 & 0.052 & {\textcolor{red} {\textbf{0.045}}} & {\textcolor{red} {\textbf{0.045}}} \\ \hline
\multirow{4}{*}{\rotatebox{90}{NLPR}~\rotatebox{90}{~\cite{peng2014nlpr}}}    & $F_{\beta}\uparrow$ & 0.687 & 0.752 & 0.841 & 0.816 & 0.841 & 0.876 & 0.888 & 0.884 & 0.904 &0.891 & {\textcolor{blue} {\textbf{0.916}}}  & {\textcolor{red} {\textbf{0.924}}} \\
                         & $S_{\alpha}\uparrow$ & 0.724 & 0.769 & 0.860 & 0.799 & 0.856 & 0.886 & 0.898 & 0.884 & 0.906 & 0.889 & {\textcolor{blue} {\textbf{0.915}}}  & {\textcolor{red} {\textbf{0.923}}} \\
                         & $E_{\theta}\uparrow$ & 0.786 & 0.840 & 0.869 & 0.884 & 0.872 & 0.926 & 0.942 & 0.920 & 0.934 & 0.937 & {\textcolor{red} {\textbf{0.949}}}  & {\textcolor{blue} {\textbf{0.942}}} \\
                         & $M\,\downarrow$ & 0.115 & 0.110 & 0.056 & 0.058 & 0.059 & 0.041 & 0.031 & 0.038 & 0.034 & 0.031 & {\textcolor{red} {\textbf{0.028}}} & {\textcolor{blue} {\textbf{0.030}}} \\ \hline
\multirow{4}{*}{\rotatebox{90}{DES}~\rotatebox{90}{~\cite{cheng2014des}}}    & $F_{\beta}\uparrow$ & 0.651 & 0.625 & 0.865 & 0.775 & 0.839 & 0.853 & 0.906 & 0.882 & 0.917 & 0.897 & {\textcolor{blue} {\textbf{0.928}}}  & {\textcolor{red} {\textbf{0.933}}} \\
                         & $S_{\alpha}\uparrow$ & 0.709 &0.685 & 0.863 & 0.770 &0.848 & 0.858 & 0.899 & 0.872 &0.904 & 0.883 & {\textcolor{blue} {\textbf{0.924}}}  & {\textcolor{red} {\textbf{0.926}}} \\
                         & $E_{\theta}\uparrow$ & 0.810 & 0.806 & 0.911 & 0.874 & 0.904 & 0.919 & 0.944 & 0.927 & 0.956 & 0.918 & {\textcolor{red}{\textbf{0.968}}}  & {\textcolor{blue} {\textbf{0.964}}} \\
                         & $M\,\downarrow$ & 0.120 & 0.131 & 0.055 & 0.068 & 0.065 & 0.046 & 0.030 & 0.038 & 0.030 & 0.030 & {\textcolor{red} {\textbf{0.023}}}  & {\textcolor{blue} {\textbf{0.027}}} \\ \hline
\multirow{4}{*}{\rotatebox{90}{LFSD}~\rotatebox{90}{~\cite{li2014lfsd}}}    & $F_{\beta}\uparrow$ & 0.680 & 0.854 & 0.815 & 0.780 & 0.813 & 0.827 & 0.872 & 0.850 & 0.849 & 0.858 & {\textcolor{blue} {\textbf{0.871}}}  & {\textcolor{red} {\textbf{0.894}}} \\
                         & $S_{\alpha}\uparrow$ & 0.658 & 0.786 & 0.796 & 0.738 &0.787 & 0.801 & 0.847 & 0.828 & 0.832 & 0.833 & {\textcolor{blue} {\textbf{0.849}}}  & {\textcolor{red} {\textbf{0.874}}} \\
                         & $E_{\theta}\uparrow$ & 0.737 & 0.841 & 0.851 & 0.810 & 0.840 & 0.851 & 0.899 & 0.867 & 0.860 & 0.875 & {\textcolor{blue} {\textbf{0.881}}}  & {\textcolor{red} {\textbf{0.907}}} \\
                         & $M\,\downarrow$ & 0.199 & 0.142 & 0.120 & 0.133 & 0.132 & 0.111 & 0.076 & 0.088 & 0.099 & 0.077 & {\textcolor{blue} {\textbf{0.079}}} & {\textcolor{red} {\textbf{0.072}}} \\ \hline
\multirow{4}{*}{\rotatebox{90}{SIP}~\rotatebox{90}{~\cite{DBLP:journals/D3/abs-1907-06781}}}    & $F_{\beta}\uparrow$ & 0.544 & 0.704 & 0.720 & 0.756 & 0.840 & 0.851 & 0.847 & 0.870 & 0.882 & 0.855 & {\textcolor{blue} {\textbf{0.892}}}  & {\textcolor{red} {\textbf{0.913}}} \\
                     & $S_{\alpha}\uparrow$ & 0.595 & 0.653 & 0.716 & 0.720 & 0.833 & 0.835 & 0.800 & 0.850 & 0.864 & 0.828 & {\textcolor{blue} {\textbf{0.875}}}  & {\textcolor{red} {\textbf{0.896}}} \\
                         & $E_{\theta}\uparrow$ & 0.722 & 0.794 & 0.824 &0.815 & 0.886 & 0.894 & 0.858 &  0.899 & 0.903 & 0.890 & {\textcolor{blue} {\textbf{0.915}}} & {\textcolor{red} {\textbf{0.923}}} \\
                         & $M\,\downarrow$ & 0.224 & 0.185 & 0.139 & 0.118 & 0.086 & 0.075 & 0.088 & 0.064 & 0.063 & 0.070 & {\textcolor{blue} {\textbf{0.054}}} & {\textcolor{red} {\textbf{0.051}}} \\ \hline

\hline
\end{tabular}
}
\end{table*}

\subsection{Datasets}
We conduct our experiments on five public challenging RGB-D datasets. NJUD~\cite{ju2015njud} contains 1985 stereo image pairs, which are captured by a Fuji stereo camera from the internet, 3D movies and photographs. NLPR~\cite{peng2014nlpr} contains 1000 image pairs, including outdoor and indoor locations which are captured by a standard Microsoft Kinect. LFSD~\cite{li2014lfsd} and RGBD135~\cite{cheng2014des} include 100 light fields images and 135 images, respectively. SIP~\cite{DBLP:journals/D3/abs-1907-06781} is a large-scale dataset which includes 929 high-resolution images. This dataset captures multiple persons in diverse scenes by a real smartphone.
We adopt the same training dataset as in~\cite{piao2019dmra}. This composited dataset contains 800 samples from DUT-RGBD~\cite{piao2019dmra}, 1485 samples from NJUD and 700 samples from NLPR. 

\subsection{Evaluation Metrics}
We adopt five widely used evaluation metrics which can represent the quality of models, namely, the Precision-Recall (PR) Curve, the $F$-measure score ($F_{\beta}$), the Mean Absolute Error ($M$), the $S$-measure ($S_{\alpha}$) and the $E$-measure ($E_{\theta}$). The $F$-measure score indicates the standard overall performance which is computed as a function of precision and recall:
\begin{equation}
    \centering
    F_{\beta }=\dfrac {\left( 1+\beta ^{2}\right) \cdot \text{P}\cdot \text{R}}{\beta ^{2}\cdot \text{P}+\text{R}}
\end{equation}
where $\beta^{2}$ is set to 0.3 to emphasize the precision~\cite{achanta2009frequencybeta2} as default, and $P$ and $R$ are obtained by using different thresholds from 0 to 255 to compare prediction and ground truth.
The parameter $E_{\theta}$ aims at capturing the image-level statistics and local pixel matching information and it is defined by:
\begin{equation}
    \centering
    E_{\theta} = \dfrac {1}{W\times H}\sum ^{W}_{x=1}\sum ^{H}_{y=1}\theta_{FM}\left( x,y\right)
\end{equation}
where W denotes the width and H denotes the height of a salient map and $\theta_{FM}$ represents the enhanced-alignment matrix~\cite{Fan2018Enhanced}. In addition, the $S_{\alpha}$ assesses the similarity on structural information and the $M$ indicates the similarity between prediction mask and ground truth mask.

\begin{figure*}
\centering
\begin{subfigure}[b]{0.3\textwidth}
\includegraphics[width=\textwidth]{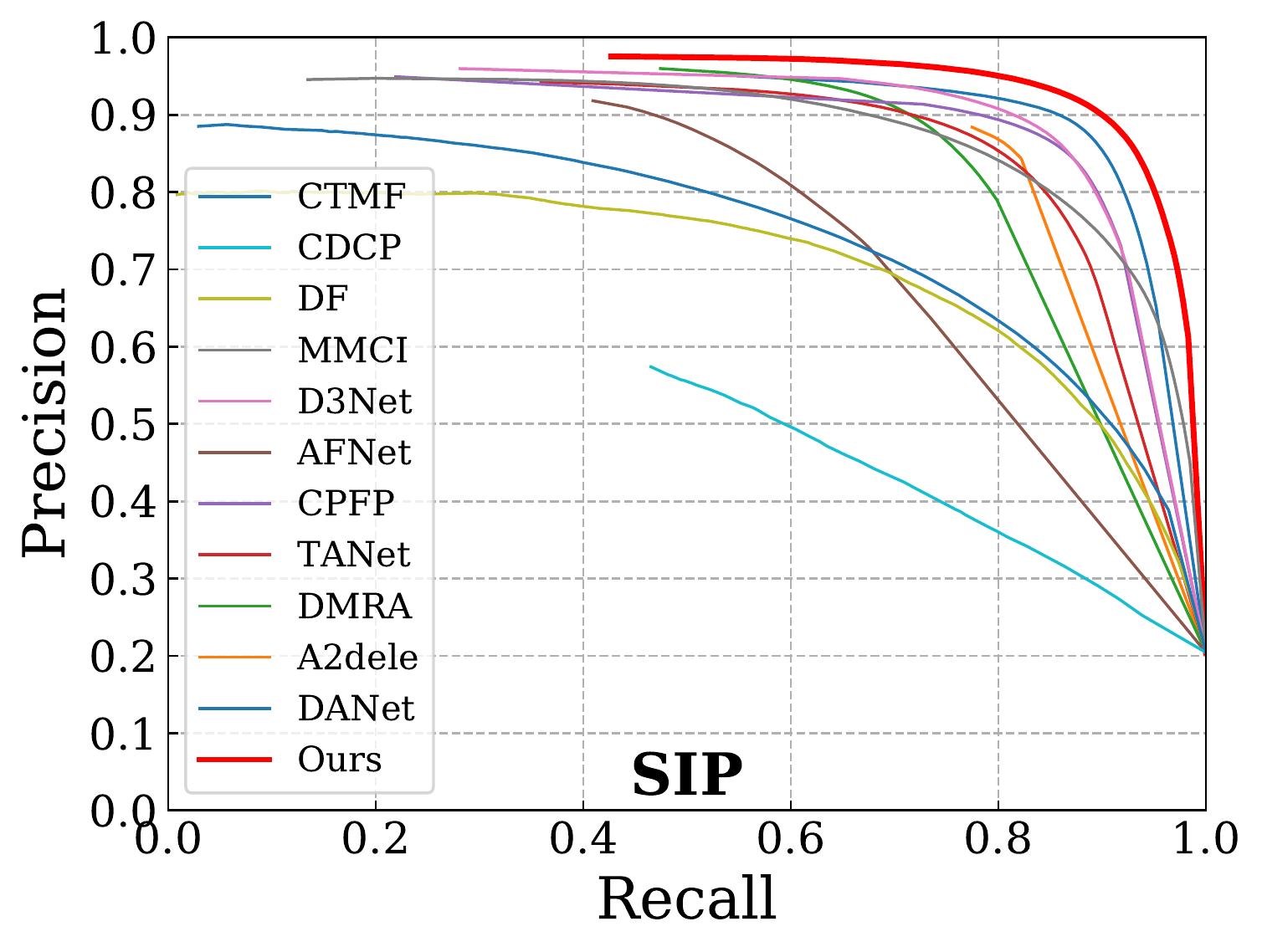}
\caption{SIP}
\end{subfigure}
\begin{subfigure}[b]{0.3\textwidth}
\includegraphics[width=\textwidth]{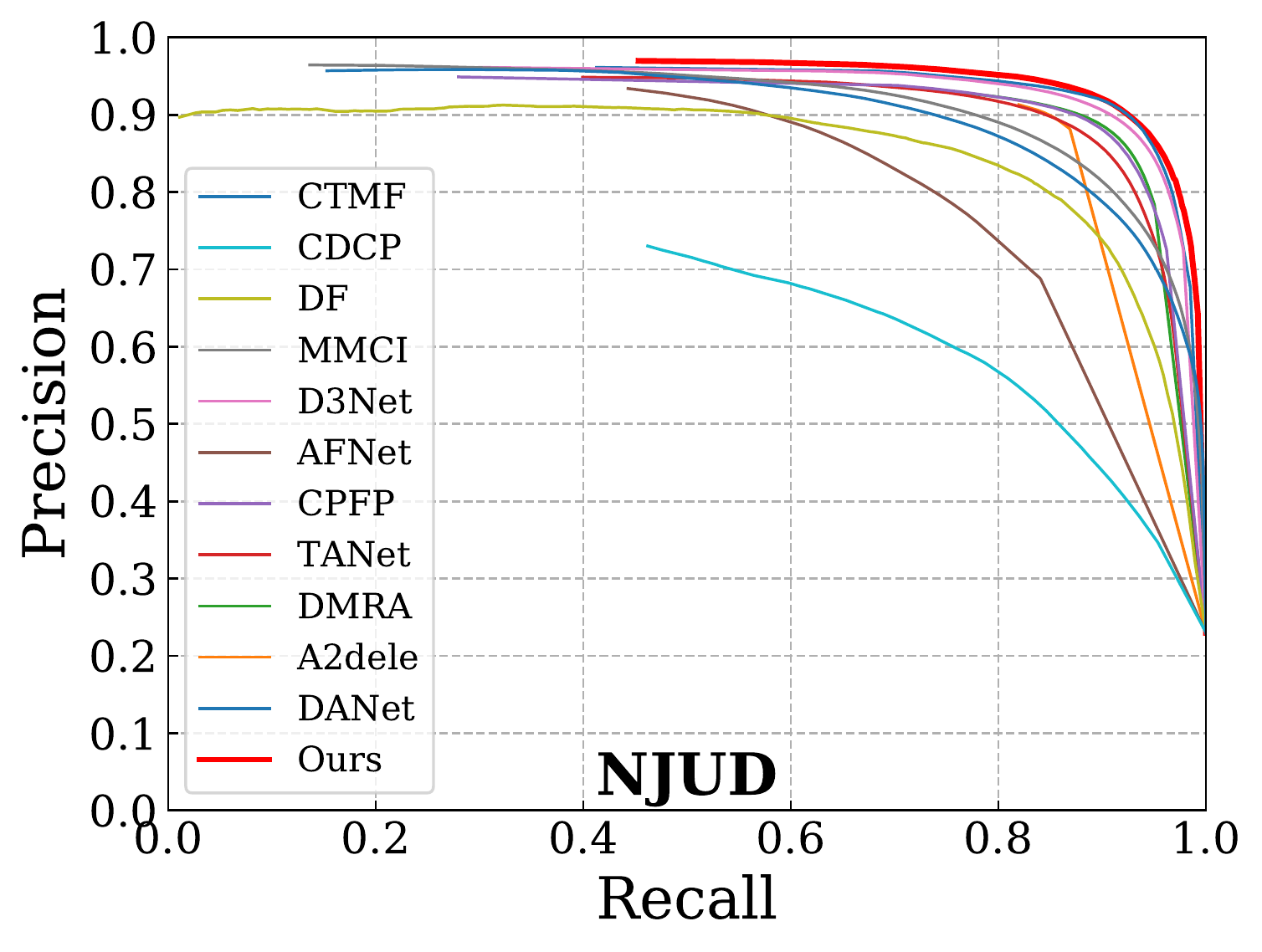}
\caption{NJU2K}
\end{subfigure}
\begin{subfigure}[b]{0.3\textwidth}
\includegraphics[width=\textwidth]{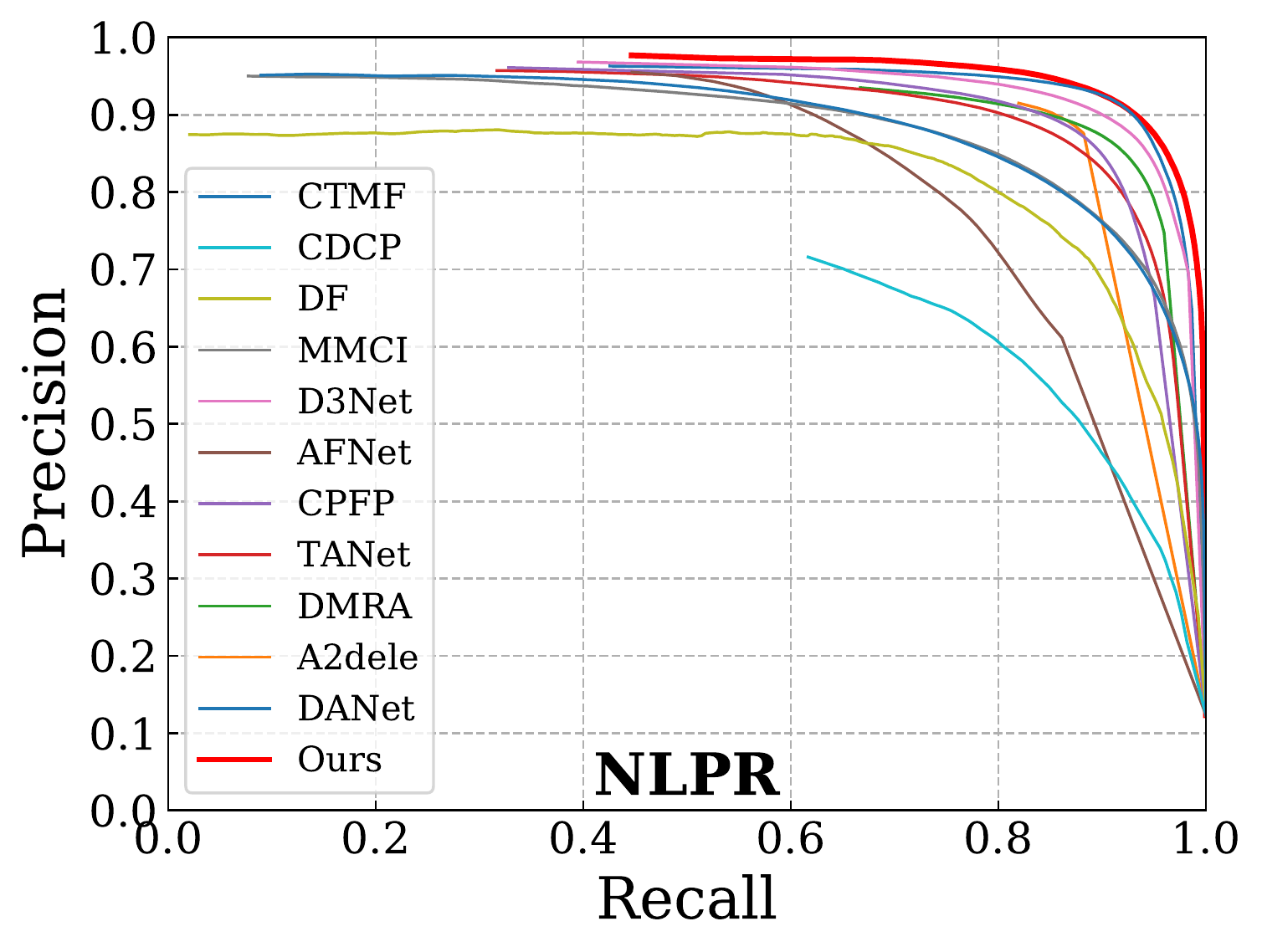}
\caption{NLPR}
\end{subfigure}

\begin{subfigure}[b]{0.35\textwidth}
\includegraphics[width=\textwidth]{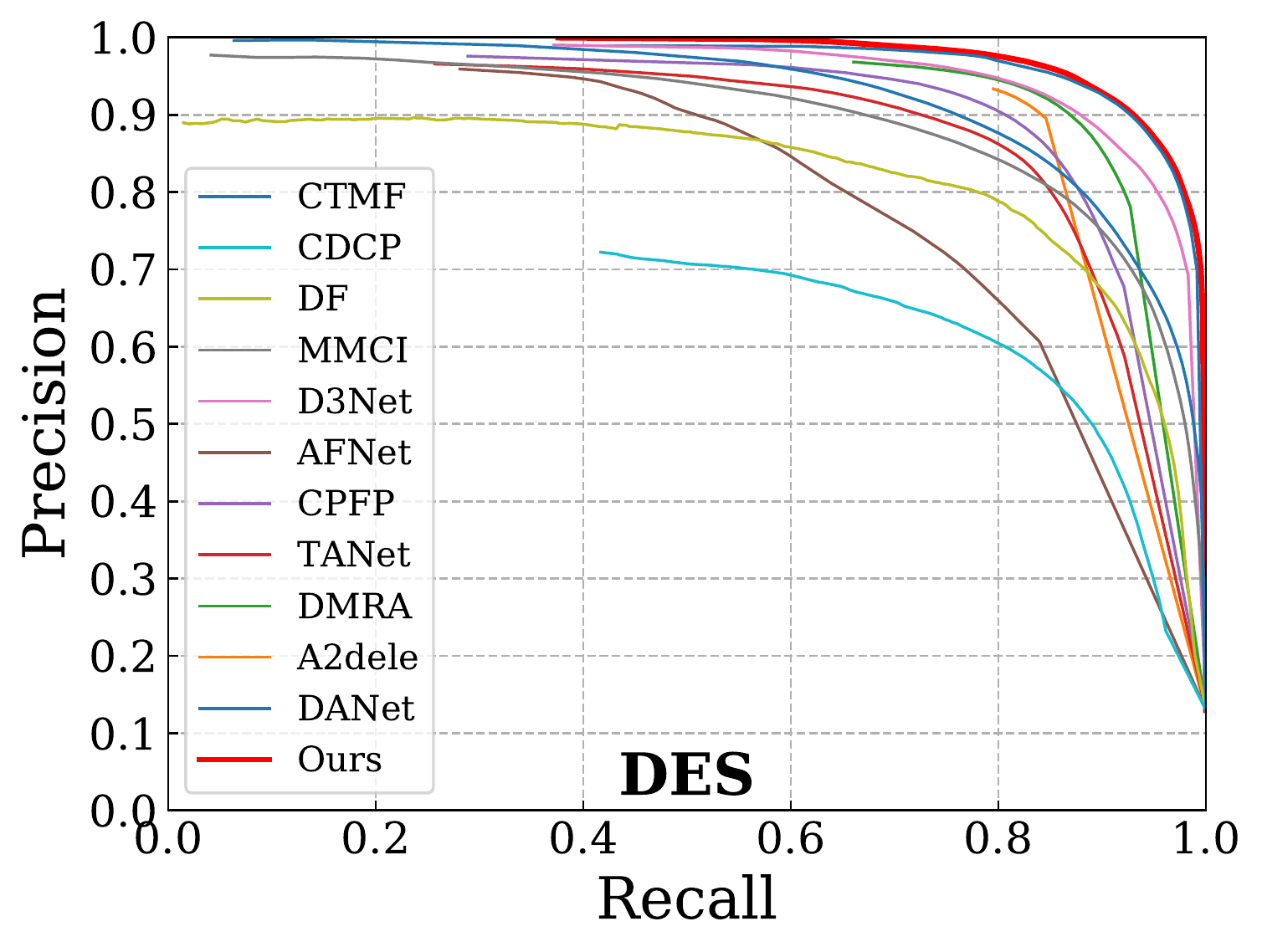}
\caption{DES}
\end{subfigure}
\begin{subfigure}[b]{0.35\textwidth}
\includegraphics[width=\textwidth]{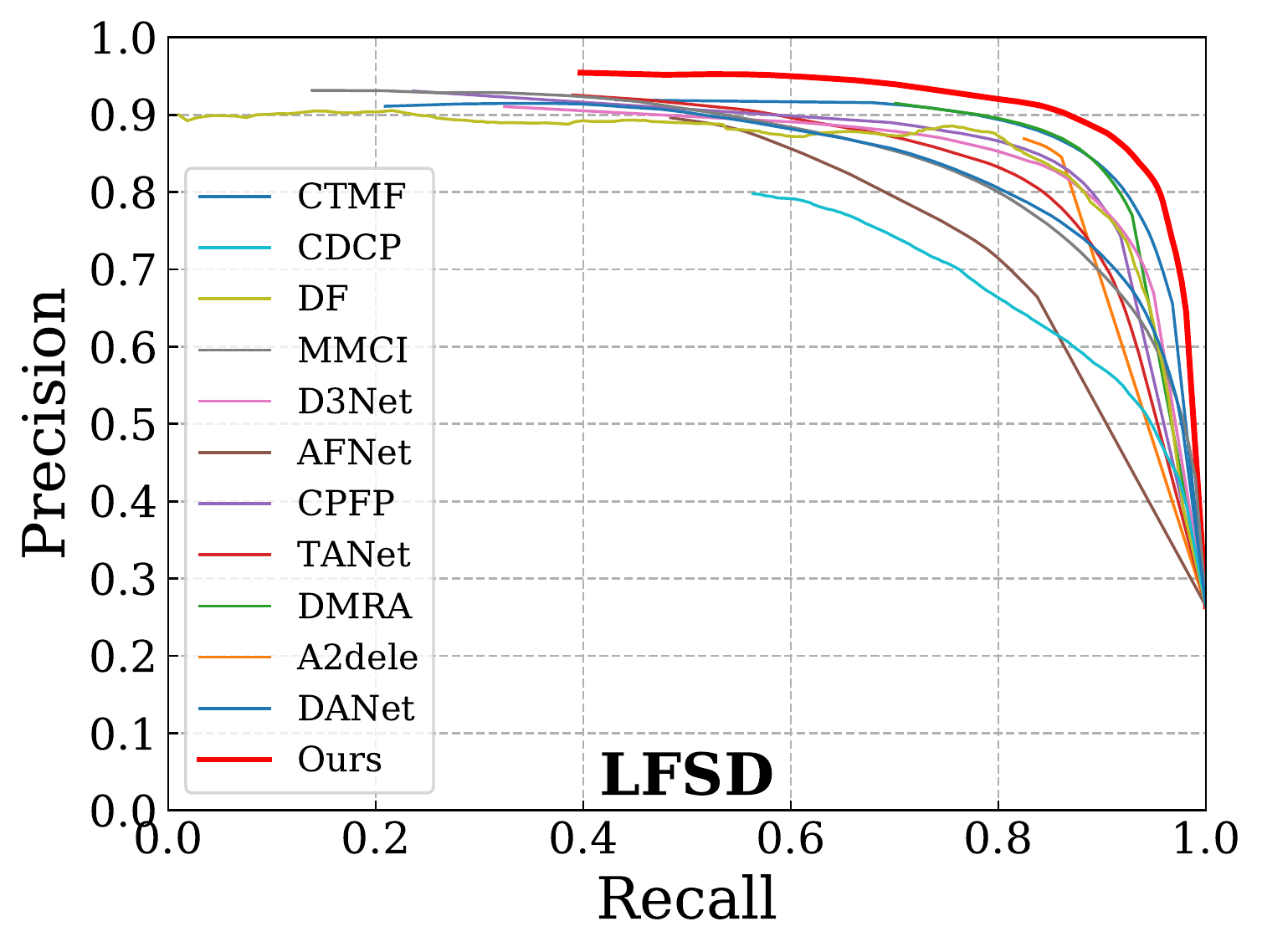}
\caption{LFSD}
\end{subfigure}
\caption{Precision-Recall curves on five public RGB-D salient object detection datasets.}
\label{pr_curve}
\end{figure*}

\begin{figure*}[t!]
\centering
  \includegraphics[width=0.9\textwidth]{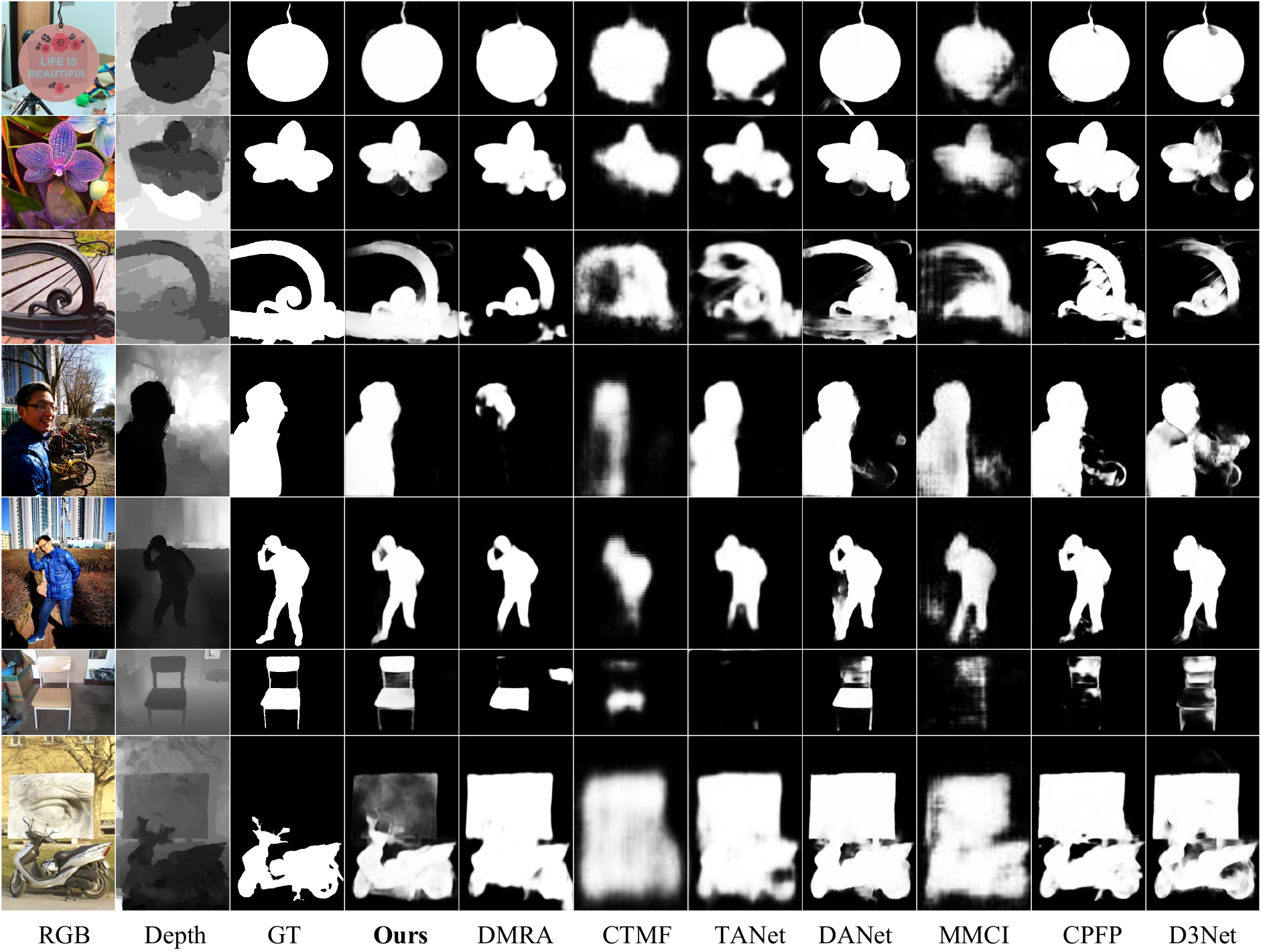}
  \caption{Visual comparisons with state-of-the-arts in different scenarios, including complicated background, low-quality depth image, low contrast between foreground and background in RGB and depth and misleading objects.}
  \label{quanlitative}
\end{figure*}

\subsection{Implementation Details}
Our model is implemented using Pytorch Toolbox and trained on a P100 GPU for 40 epochs with mini-batch size 4. We use two identical VGG19 as feature extraction backbones for RGB stream and depth stream. Both RGB and depth images are resized to 256x256. To avoid overfitting, simple flipping and rotating are adopted to augment the training dataset. The initial learning rate is set to 1e-3 and using poly policy~\cite{liu2015polypolicy}. We adopt a 0.0005 weight decay for the stochastic gradient descent (SGD) with a momentum of 0.9 and binary cross entropy loss function for supervision.

\subsection{Comparison with State-of-the-art Results}
We compare our model with 11 state-of-the-art methods on five challenging benchmark datasets, including CDCP~\cite{zhu2017innovative}, DF~\cite{qu2017df}, AFNet~\cite{wang2019afnet}, CPFP~\cite{zhao2019contrast}, MMCI~\cite{chen2019mmci}, TANet~\cite{chen2019tanet}, CTMF~\cite{han2017ctmf}, D3Net~\cite{DBLP:journals/D3/abs-1907-06781}, DMRA~\cite{piao2019dmra}, A2delde~\cite{piao2020a2dele} and DANet~\cite{zhao2020single}. For fair comparisons, we directly use the saliency maps released by the authors or the evaluation results pre-computed by the authors.

\noindent\textbf{Quantitative Evaluation.}
Table \ref{table:qua} and Figure \ref{pr_curve} show the quantitative results in comparison to 11 state-of-the-art methods in terms of four evaluation metrics on five challenging datasets. More specifically, it is observed in Table \ref{table:qua} that our proposed method outperforms all other methods in terms of maximum $F_{\beta}$ and $S_{\alpha}$ across five datasets. Especially, our method outperforms all other methods by a large margin in four metrics on SIP, which is a large-scale dataset with relatively complicated scenes. Figure \ref{pr_curve} shows the comparison results on PR curve, which computes the precision and recall between the binary mask and ground truth. It can be clearly observed that our method which is represented by the red line outperforms other state-of-the-art methods. This observation illustrates the better performance of our model at different thresholds.

\noindent\textbf{Qualitative Evaluation.}
Figure \ref{quanlitative} illustrates the visual comparison with other state-of-the-art methods. Generally, the results from our model are more similar to the ground truth in different scenarios, which shows the superiority of our method. For instance, in low contrast scenes(row 3 and row 5), our method can accurately detect the salient object with more semantic details. In addition, our method shows an accurate prediction in a challenging situation(row 7). More specifically, due to the board in the background, which is located in the center of the scene, the board is viewed as the salient object of interest in other methods and it is associated with high confidence values in the predicted maps. Compared to them, there are less non-salient parts shown in our result. Furthermore, in complex environments, even the depth maps have high contrast between the foreground and background(row 1, row 2 and row 5), the existing methods still detect undesired parts due to the complicated background in the RGB maps. Compared to other methods, our method can effectively filter the non-salient parts and focus more on the salient objects in the complex environments. 

\begin{figure}[t!]
  \centering
  \includegraphics[width=\columnwidth]{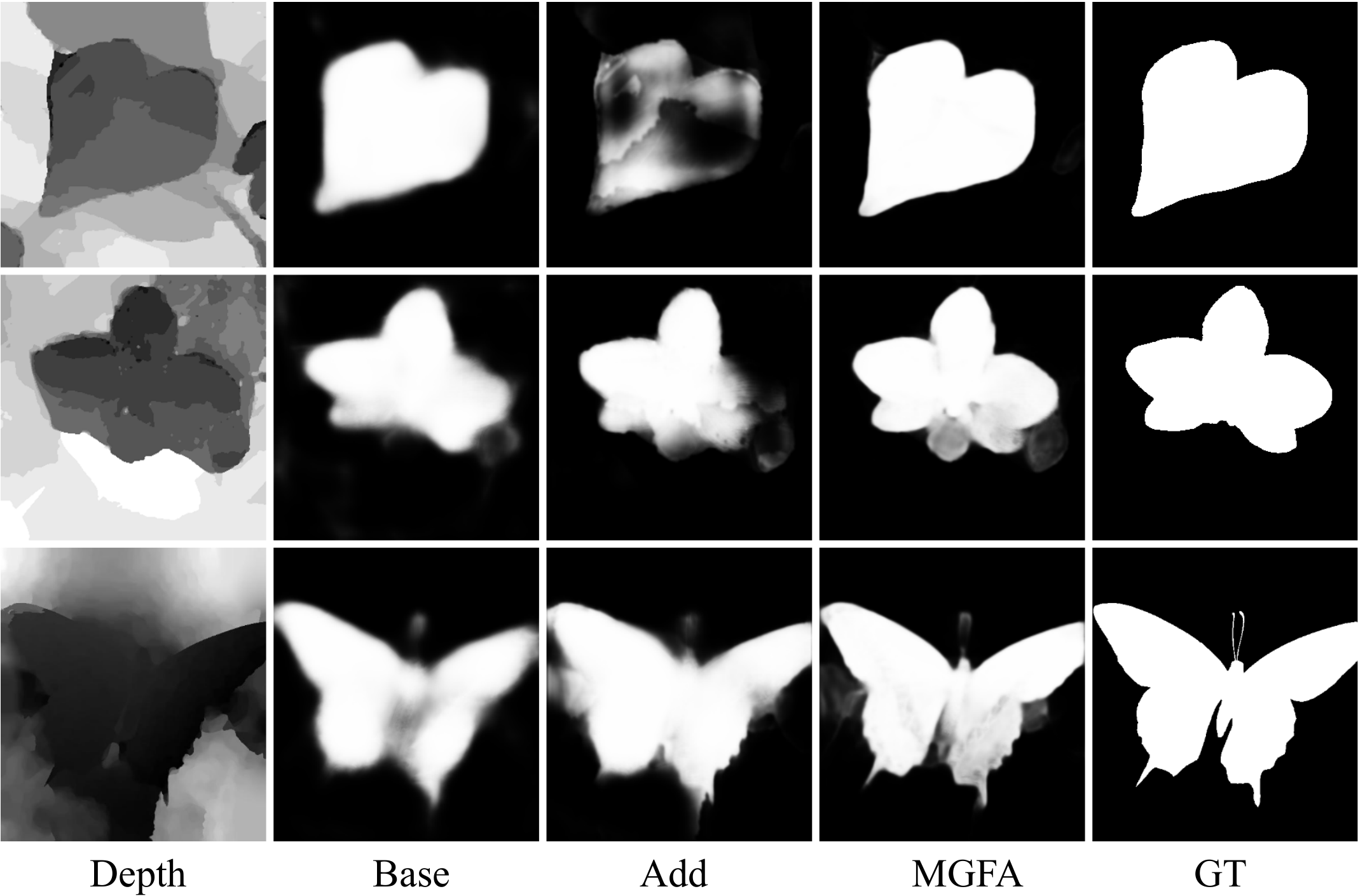}
\caption{Effectiveness of MGFA. $Base$ represents the FPN baseline, $Add$ means combining two-modal features by summation in the same two-stream framework.}
\label{mgfa_effect}
\end{figure}

\subsection{Ablation Studies}
In this section, we conduct a series of experiments to verify the effectiveness of the proposed methods. The baseline is an FPN network which is built on a VGG19 backbone.

\noindent\textbf{Mask-Guided Feature Aggregation.} Figure \ref{mgfa_effect} illustrates the effectiveness of MGFA. $Add$ and $MGFA$ represent two multi-scale fusion strategies in the same two-stream network. It can be observed in row 1 and row 2 that directly adding the RGB and depth features can impede the detection performance compared to the results of $Base$. The bottom row shows the prediction results with a low-contrast depth map. Although simple addition between two-modal features can improve the semantic features, irrelevant information is also introduced. Compared to direct combination in the multi-scale fusion framework, MGFA shows a better performance on cross-modal feature aggregation. 

\begin{table}
\caption{Ablation analyses on NLPR, LFSD and SIP. Base means a FPN network built on a VGG19 backbone. MGFA and MGRM are two proposed modules. ASPP is an Atrous Spatial Pyramid Pooling module to enlarge the receptive field.}
\label{table:ablation}

\resizebox{\columnwidth}{!}{
\begin{tabular}{cc|ccc|c}
\hline
\hline
\multirow{2}{*}{} & Metric & Base & +MGFA  & +ASPP    & +MGRM  \\
                         \hline
\multirow{4}{*}{\rotatebox{90}{NLPR}~\rotatebox{90}{~\cite{peng2014nlpr}}}    & $F_{\beta}\uparrow$  & 0.865 & 0.917 &0.920  & {\textcolor{black} {\textbf{0.924}}} \\
                         & $S_{\alpha}\uparrow$  & 0.877 & 0.917 & 0.922  & {\textcolor{black} {\textbf{0.923}}} \\
                         & $E_{\theta}\uparrow$  & 0.893 & 0.934 & 0.940  & {\textcolor{black} {\textbf{0.942}}} \\
                         & $M\,\downarrow$  & 0.051 & 0.033 & 0.031 & {\textcolor{black} {\textbf{0.030}}} \\ \hline

\multirow{4}{*}{\rotatebox{90}{LFSD}~\rotatebox{90}{~\cite{li2014lfsd}}}    & $F_{\beta}\uparrow$ & 0.856 & 0.865 & 0.878  & {\textcolor{black} {\textbf{0.894}}} \\
                         & $S_{\alpha}\uparrow$  & 0.825 & 0.850 & 0.863 & {\textcolor{black} {\textbf{0.874}}} \\
                         & $E_{\theta}\uparrow$  & 0.858 & 0.894 & 0.894  & {\textcolor{black} {\textbf{0.907}}} \\
                         & $M\,\downarrow$  & 0.112 & 0.082 & 0.081 & {\textcolor{black} {\textbf{0.072}}} \\ \hline
\multirow{4}{*}{\rotatebox{90}{SIP}~\rotatebox{90}{~\cite{DBLP:journals/D3/abs-1907-06781}}}    & $F_{\beta}\uparrow$ & 0.821 & 0.910 & 0.911  & {\textcolor{black} {\textbf{0.913}}} \\
                     & $S_{\alpha}\uparrow$ & 0.818 & 0.890 & 0.891  & {\textcolor{black} {\textbf{0.896}}} \\
                         & $E_{\theta}\uparrow$  &  0.862 & 0.918 & 0.920 & {\textcolor{black} {\textbf{0.923}}} \\
                         & $M\,\downarrow$ & 0.101 & 0.054 & 0.055 & {\textcolor{black} {\textbf{0.051}}} \\ \hline

\hline

\end{tabular}
}
\end{table}

\begin{figure}
  \includegraphics[width=\columnwidth]{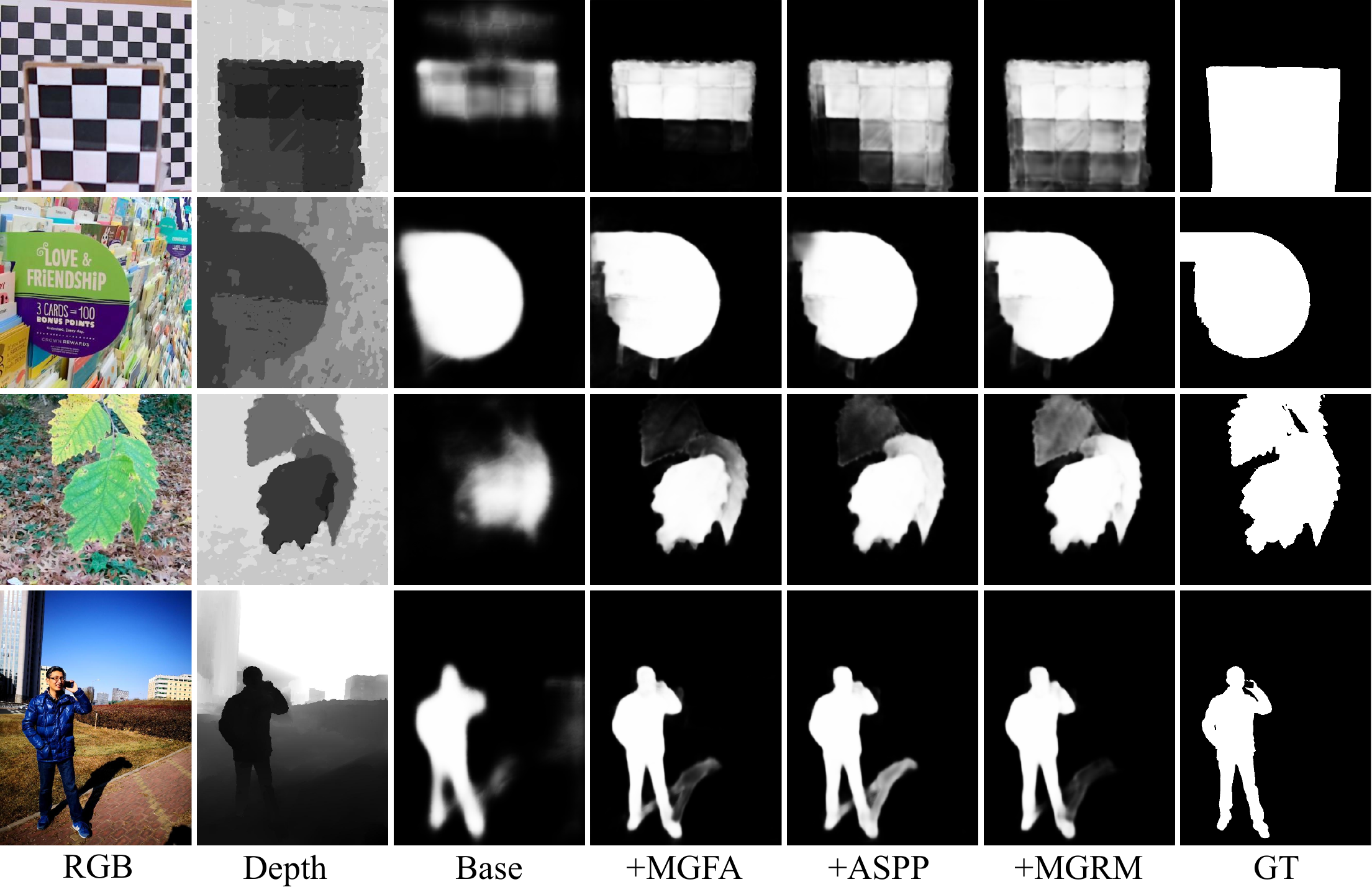}
  \caption{Visual comparisons of the proposed modules. For each experiment we employ the corresponding modules on the framework. Specially, MGFA and MGRM are employed on five scales and the ASPP module is only applied at the end of the RGB stream.}
  \label{ablation_qualitative}
\end{figure}

Additionally, Table \ref{table:ablation} shows the quantitative results of MGFA in the second column. It is demonstrated that the MGFA improves the baseline in all metrics across three datasets. Especially, the MGFA boosts the performance by a large margin on SIP, providing gains of $8.9\%$, $7.2\%$, $5.6\%$ and $4.7\%$ in terms of the maximum $F_{\beta}$, $S_{\alpha}$, $E_{\theta}$ and $M$ respectively. On the other hand, Figure \ref{ablation_qualitative} illustrates the visual results after adopting MGFA in the fourth column. More specifically, row 1 shows a salient object in a particular scene where foreground and background possess similar structures. Row 2 and row 3 show complex background in RGB maps. Row 4 shows a depth map in the case of low contrast between the salient object under consideration and background. Compared to the predictions of baseline, using MGFA could alleviate the impact of low-quality depth maps and effectively exploit their semantic information. Furthermore, due to the smaller effective receptive field\cite{luo2017understandingreceptive}, we add an ASPP module at the end of RGB stream to further enlarge the receptive field for large-scale objects.

\noindent\textbf{Mask-Guided Refinement Module.} We employ the MGRM to build the proposed model. The effectiveness of MGRM can be demonstrated in the last column of Table \ref{table:ablation}. It can be observed that all evaluation metrics are improved by the proposed MGRM. Evidently, this module improves the overall performance and reduces the error by a large margin on LFSD, providing gains of $1.6\%$, $1.1\%$, $1.3\%$ and $0.9\%$ in terms of the maximum $F_{\beta}$, $S_{\alpha}$, $E_{\theta}$ and $M$ respectively. Furthermore, 
Figure \ref{ablation_qualitative} illustrates the effectiveness of MGRM in visual maps in the sixth column. Compared to the previous columns, it is obvious that row 1 and row 3 in the sixth column are closer to the ground truth, which demonstrates that the MGRM can effectively complement high-level semantic features. On the other hand, row 2 and row 4 show a better performance in preserving details, especially in the bottom area where the contrast is quite low in the depth maps. These results indicate that the MGRM can further refine the saliency maps.

\noindent\textbf{Hyperparameters Setting.} We conduct experiments on a different number of scales to demonstrate the proposed progressive multi-scale fusion strategy. Table \ref{tab:scales} shows the results on three different scales across three datasets. Specifically, $s$ represents the number of scales. For instance, $s=3$ indicates that we only combine the $\left\{f^3_{r},f^4_{r},f^5_{r}\right\}$ and $\left\{f^3_{d},f^4_{d},f^5_{d}\right\}$ by employing three MGFAs and MGRMs. It is seen that using only three scales can obtain a large gain on SIP in terms of all metrics, namely, $8.1\%$ on $F_{\beta}$, $6.7\%$ on $S_{\alpha}$, $5.5\%$ on $E_{\theta}$ and $4.4\%$ on $M$. We have shown that the semantic features are mainly contained in the deep layers, therefore, three-scale fusion can achieve accurate performance. Furthermore, the experimental results also indicate that increasing the number of scales could lead a better overall performance. In our experiment, $s$ is set to 5 for the progressive multi-scale fusion.

\begin{table}
\caption{Quantitative comparison with different scales on NLPR, LFSD and SIP.}
\label{tab:scales}

\resizebox{\columnwidth}{!}{
\begin{tabular}{cc|ccc|c}
\hline
\hline
\multirow{2}{*}{} & Metric & Base & s=3  & s=4    & s=5  \\
                         \hline
\multirow{4}{*}{\rotatebox{90}{NLPR}~\rotatebox{90}{~\cite{peng2014nlpr}}}    & $F_{\beta}\uparrow$  & 0.865 & 0.907 &0.921  & {\textcolor{black} {\textbf{0.924}}} \\
                         & $S_{\alpha}\uparrow$  & 0.877 & 0.914 & 0.921  & {\textcolor{black} {\textbf{0.923}}} \\
                         & $E_{\theta}\uparrow$  & 0.893 & 0.935 & 0.941  & {\textcolor{black} {\textbf{0.942}}} \\
                         & $M\,\downarrow$  & 0.051 & 0.033 & 0.030 & {\textcolor{black} {\textbf{0.030}}} \\ \hline

\multirow{4}{*}{\rotatebox{90}{LFSD}~\rotatebox{90}{~\cite{li2014lfsd}}}    & $F_{\beta}\uparrow$ & 0.856 & 0.891 & 0.891  & {\textcolor{black} {\textbf{0.894}}} \\
                         & $S_{\alpha}\uparrow$  & 0.825 & 0.865 & 0.872 & {\textcolor{black} {\textbf{0.874}}} \\
                         & $E_{\theta}\uparrow$  & 0.858 & 0.895 & 0.901  & {\textcolor{black} {\textbf{0.907}}} \\
                         & $M\,\downarrow$  & 0.112 & 0.079 & 0.076 & {\textcolor{black} {\textbf{0.072}}} \\ \hline

\multirow{4}{*}{\rotatebox{90}{SIP}~\rotatebox{90}{~\cite{DBLP:journals/D3/abs-1907-06781}}}    & $F_{\beta}\uparrow$ & 0.821 & 0.902 & 0.907  & {\textcolor{black} {\textbf{0.913}}} \\
                     & $S_{\alpha}\uparrow$ & 0.818 & 0.885 & 0.890  & {\textcolor{black} {\textbf{0.896}}} \\
                         & $E_{\theta}\uparrow$  &  0.862 & 0.917 & 0.918 & {\textcolor{black} {\textbf{0.923}}} \\
                         & $M\,\downarrow$ & 0.101 & 0.057 & 0.055 & {\textcolor{black} {\textbf{0.051}}} \\ \hline

\hline

\end{tabular}
}
\end{table}

\section{CONCLUSIONS}
In this paper, we aim to improve the prediction accuracy towards RGB-D based SOD. We design a progressive multi-scale fusion architecture which is built on a two-stream framework. We propose a novel cross-modal feature aggregation module which could effectively introduce the depth information and alleviate the noise from low-quality depth maps. Furthermore, we propose a mask-guided refinement module to refine the salient maps and improve the performance. This module could complement high-level semantic features and filter the noise from multi-scale features due to the progressive mask-guided strategy. Both quantitative and qualitative experimental results demonstrate the effectiveness of the proposed modules verifying that the proposed framework achieves competitive results on five public challenging benchmarks.

{\small
\bibliographystyle{ieee}
\bibliography{egpaper_final}
}

\end{document}